\documentclass[journal]{IEEEtran}
\usepackage{latexsym}
\usepackage{graphicx}
\usepackage{amsfonts,amssymb,amsmath}
\usepackage{hyperref}
\usepackage[compact]{titlesec}         
\titlespacing{\section}{3pt}{3pt}{3pt} 
\AtBeginDocument{
  \setlength\abovedisplayskip{2pt}
  \setlength\belowdisplayskip{2pt}}
\def\BibTeX{{\rm B\kern-.05em{\sc i\kern-.025em b}\kern-.08em T\kern-.1667em\lower.7ex\hbox{E}\kern-.125emX}}
\begin{document}

\title{From DNNs to GANs: Review of efficient hardware architectures for deep learning}

\author{Gaurab Bhattacharya
\thanks{G. Bhattacharya is with the School of Electrical Sciences, Indian Institute of Technology Bhubaneswar, Pin: 752050, India. \protect E-mail: gb14@iitbbs.ac.in.}}

\maketitle

\begin{abstract}
In recent times, the trend in very large scale integration (VLSI) industry is multi-dimensional, e.g. reduction of energy consumption, occupancy of less space, precise result, less power dissipation, faster response etc. To meet these needs, the hardware architecture should be reliable and robust to these problems. Recently, neural network and deep learning has been started to impact the present research paradigm significantly which consists of parameters in the order of millions, nonlinear function for activation, convolutional operation for feature extraction, softmax regression for classification, generative adversarial networks, etc. These operations involve huge calculation and memory overhead. Presently available DSP processors are incapable of performing these operations and they most face the problems e.g. memory overhead, performance drop and compromised accuracy. Moreover, if a huge silicon area is powered to accelerate the operation using parallel computation, the IC’s will be having significant chance of burning out due to the considerable generation of heat. Hence, novel dark silicon constraint is developed to reduce the heat dissipation without sacrificing the accuracy. Similarly, different algorithms have been adapted to design a DSP processor compatible for fast performance in neural network, activation function, convolutional neural network and generative adversarial network. In this review, we illustrate the recent developments in hardware for accelerating the efficient implementation of deep learning networks with enhanced performance. The techniques investigated in this review are expected to direct future research challenges of hardware optimization for high-performance computations.
\end{abstract}


\section{INTRODUCTION}

\IEEEPARstart{T}{he} technological imperialism in the frontiers of the miniaturization of computing devices has dramatically reduced the size of the computers from room size giants to pocket size nomadic devices. Despite decrease in size, their computational power has increased manifold. This evolution in size and speed of computers has resulted in their widespread adoption by society as nomadic devices or personal digital assistants (PDAs). The importance of nomadic computing devices is profound, and it is being surveyed that the mobile phone population in the world has reached six billion users among a world population of seven billion people. However, this wonderful trend of engaging people into better connectivity is undoubtedly not limited to mobile phones. Portable computing devices are penetrating into society by proving their usefulness in the field of bio-informatics, security, health monitoring, navigation, music and imagining etc. The portability, battery efficiency, sustainability and the precised and fast performance are some of the challenges faced by the VLSI research community, that dictates that performance should be improved 1000$\times$ with power budget increment of only 40\% and no increase in design team size. 

\par Similar to this field, the high-performance computing (HPC) community also has similar challenges going to perform more operations in a decade. And even if these computers are driven by mains and not battery they have equal, if not more, stiff challenge of improving the performance/power ratio because of cooling and the price of electricity and environmental concerns. To achieve the needs due to the latest technological advancement, high performance and low power computing community is facing the challenge of growth rate gap between application complexity, VLSI technology, design technology and battery technology.

\par Hence, in recent years, the development of Systems on Chip (SoCs) is facing increasing challenges, which are starting to expose the limitations of current design philosophies. In spite of many conservative measures, often at the expense of computational efficiency, realising a functional SoC has come to require tens of Millions USDs. Such a tendency threatens to hinder innovation and keeps away applications that require the efficiency of ASIC. Among them, a prime example are Artificial Neural Networks (ANNs), which have become one of the most popular research topics and already witnessed many commercial applications in countless fields. Parallel to this development, convolutional neural network (CNN) and generative adversarial network (GAN) attribute to the demand of high-performance computing architecture. The increasing complexity of these networks make them quite demanding in terms of computational power, so the inefficient approach of modern SoCs is quickly becoming unsuited for them. Implementing popular algorithms such as ANNs on a state-of-the-art platform is a good opportunity to prove its advantages and this is precisely the rationale behind this review.

\par This paper is organized as follows: Section II describes the motivation and recent developments to obtain high-performance complex hardware architecture, Section III illustrates the SiLaGO solution and its internal architecture, Section IV describes the developments in hardware to perform artificial neural network (ANN), Section V presents the state-of-the-art solutions for softmax activation function implementation, Section VI describes the latest architectures for hardware implementation of convolutional neural network (CNN), Section VII gives an overview of generative adversarial network (GAN) implementation methodologies using hardware and Section VIII discusses the future scope in this research during conclusion. 

\section{Motivation and Recent Developments}
Technology scaling is a continuous process which tries to accumulate more number of transistors in a single chip with low power consumption. However, the design becomes more complex and the power requirement increases which outstrip the benefit of technology scaling. This phenomenon has been mentioned in the Hotchips tutorial by Rabaey and it is referred to as ``Shannon beats Moore'' \cite{paper1}. This observation, which is true for for different domains of application can be called upon as the architecture efficacy gap. There are different innovations happened in this domain to reduce these problems; e.g. virtual memory, pipelining, multi core processor, instruction and thread-level parallelism etc. These methods are aimed for providing better engineering and computational efficiencies and better silicon use. 

\subsection{Dark silicon constraint}
During recent decades, the focus of efforts of integrated circuit manufacturers were upon Moore’s Law where increment in density of transistor happens at constant cost. During the same period, Dennard’s Law also held: which tells the power consumption goes down with the dimensions of a device go down, and less cost and power will be incurred by smaller transistors which can ran faster. Previously, number of transistors in an IC was limited; hence adding more transistors increased the power consumption significantly. Heat is different for every watt of power consumed. Unfortunately, the power each transistor consumed didn't decrease at a promising rate while the number of transistors that can be cheaply put on a chip has increased. Indeed, there is a decrement in the power consumption of transistor, but it decreased in lesser pace than a transistor shrunk \cite{paper9}. Hence to control heat dissipation and prevent the IC from burning out, all the transistors in an IC cannot be powered together. This is termed as ``Dark silicon problem''.

\par This situation resulted in an era where Dennard scaling fails which is known as ``dark silicon''. If transistors doubles in number, but the circuit's power budget remains as a whole the same or goes down, because of the advancement of mobile devices, then the power available for each transistor is cut in half. If we keep same threshold voltage, then we also have to reduce the number of transistors operable at one time into half. These transistors which are non-operational are referred to be as dark silicon, which is measured as the fraction of unused area of the chip. In many of today’s designs, many circuit paths will be ``dark'' at any given moment. Hence, the power consumption rate will consider power to only a fraction of a chip which results the percentage of the chip which is active with generation of each processor to be exponentially smaller \cite{paper2}.

\par However, there can be a possible solution to this problem. As we know that the dark silicon constraint prevents transistors more than a pre-defined number to turn on at a time, we need to use transistors with high silicon and computational efficiency.To tackle this, VLSI experts have suggested two solutions: Heterogeneity of processors and use of accelerators. Also, if we want to tackle the cost due to steep engineering, we have to settle on designs which are platform-based and consists of mostly general-purpose processors and interconnects associated to them \cite{paper2}.

\subsection{Limitations in researches on dark silicon constraint}

Since its inception, detailed work has been performed to combat the constraint of dark silicon. The existing research on dark silicon constraint can be subdivided into four horizontals. The first of them is the decision of the platform to be used for chips in dark silicon era. General purpose processor, undoubtedly effective are considered to be less efficient for some cases. Another horizontal deals with the power management using dynamic voltage and frequency scaling to make the power consumption below the budget. Also, finding out proper additional components has yielded better performance, or sometimes better algorithms can produce efficient chip design within a power budget. Near-threshold voltage computation is another aspect of research which improves the performance by keeping the supply voltage very low. However, it can be argued that the state-of-the-art design methodology are inadequate in three aspects, as given below:

\begin{itemize}
\item Partial Computation-centric customization: Overall power generation and consumption is significantly affected by the interconnect, generation of control statements and storage. More than 50\% power is consumed in 130 nm technology \cite{paper3}. Similarly, another cost-centric area in the chip is memory, hence it should be customizable according to the application.

\item Inefficient software-centric implementation style: The software-centric customization style uses their accelerators to keep the parts which are performance and power critical.. Hence bulk of functionality is not mapped into the accelerator which makes the system inefficient. 

\item Large engineering cost for customization: The newly found custom hardware is expensive to design. In spite of considering new technology, the engineering cost for a complex SOC is 45 million USDs in addition to five million USDs in manufacturing cost.
\end{itemize}

\subsection{Motivation for SiLaGO}
First part of design flow is a manual operation which consists of designing accelerator and SOC. The second part is logic implementation which can be automated, but can lead to costly iteration procedures which make SOC manufacturing a very complicated proposition and unless this challenge is addressed, the objective of complete customization to achieve significantly more functionality per unit current and area to deal with dark silicon constraints cannot be fulfilled. The architectural template that is used by SiLaGO is hardware centric and enables complete and dynamic customization to overcome problems.

\section{Overview of hardware accelerators}
Hardware accelerators have used application-specific integrated circuits (ASIC), field-programmable gate array (FPGA), graphic processing units (GPU), silicon large grain objects (SiLaGO), etc. to obtain better throughput for state-of-the-art deep learning algorithms. This section provides an overview of these methods adopted by hardware accelerators.

\subsection{Application-specific integrated circuit (ASIC)}
ASIC, a high-density integrated circuit is designed for customer-specific application, rather than general purpose use. Modern ASICs consist of microprocessors, variants of memory blocks, such as EPROM, flash memory, etc. altogether considering up to 100 million logic gates. There are four variants of ASICs; namely, gate arrays, structured ASICs, standard cell, full-custom, in the increasing order of complexity. Gate arrays are design to comprise basic cells with transistors and interconnects for user-specific applications by predetermining optimum configuration for logic gates. Semi-customization of multiple layers to reduce manufacturing time and cost are designed by structured ASICs. Standard cells are used to optimize the internal resources of gate arrays to create customized layers for every application. In full-custom design, inherent control over every mask layer and intricate design incorporate the most complex ASIC architecture.

\par The customized architecture and high-degree of parallelism have suited ASICs for widespread implementation of pretrained deep learning networks. Its parallel resource allocation mechanism alleviate the problem of resource utilization by dynamically utilizing memory for gradient-based operations. For state-of-the-art deep learning training and inference mechanism, tensor processing unit has been proposed made of an ASIC-built system which optimizes high-speed low-precision floating-point operations by consuming lesser power than GPU.

\subsection{Field-programmable gate array (FPGA)}
FPGAs are off-the-shelf programmable integrated circuits to provide low-cost implementation of customer-specific hardware functionalities. FPGAs consist of configurable logic block and programmable I/O cells around the device with programmable interconnections. Several digital signal processing (DSP) blocks, such as block RAM, flip-flops, look-up tables, multiply-and-accumulate, clock units, etc. are also embedded to perform memory and computation-intensive operations. Due to this design advantage, hardware accelerators have seen a widespread use of FPGAs with highly flexible models with fine-grained parallelism \cite{fpga1}. Also, acceleration in application-specific tasks, such as data streaming, parallel access of memory, stacks and queues, etc. can be performed using FPGAs with optimized performance in low-power, which reduces large-scale infrastructure cost \cite{fpga2,fpga3}. 

\par The benefits of FPGAs make them a suitable candidate for deep learning applications for low-power devices or cloud services. Also, the large degree of parallelism and implementation flexibility of FPGAs result in high execution speeds required for deep learning. Also FPGAs can create customized accelerators with multi-threading which supports inherent nature of feed-forward neural network. The reconfigurable nature of the FPGAs also attribute the deep learning framework of partial reconfiguration of weights during backpropagation. Moreover, this partial reconfiguration method can be augmented with kernel size variations in CNN and GAN applications, making FPGA suitable for a wide range of deep learning architectures.

\subsection{Graphics processing unit (GPU)}
GPU, a specialized electronic circuit is largely used for the fast manipulation and acceleration of image processing and computer graphics applications using memory alterations with multiple parallel core. A GPU consists of graphics and computer array, compression unit, graphics memory controller, video processing and power management units, bus and display interfaces. Modern GPUs performs 3-D graphics related calculations by using most of their transistors. Also, basic 2-D acceleration and DSP operations can be performed by this. 

\par Parallel threads in the order of thousands in GPUs help to accelerate training and inference of the state-of-the-art deep learning architectures by maximizing floating-point throughput. Efficient use of memory bandwidth makes GPU work faster than CPU by the use of video random access memory (VRAM) and dedicating CPU memory for other functional use. Also, traditional deep learning algorithms constitute large dataset which attributes to the large computation time and memory cost, which can be efficiently using using parallel memory allocation by GPUs. 

\subsection{The SiLaGO solution}
SiLaGO blocks are \textit{synchoros} architecture that implements small-scale architecture levels operations and are 4-5 times larger than the standard cells from boolean level. These are used as a building block which can be spatially composed by abutment, no further VLSI engineering is required \cite{paper4}. The architecture of SiLaGO blocks follow \textit{synchorocity} and are made up of thinking to minimize the inter-connecting wires. The blocks are made in such a way that the sub-components are placed in a place with minimal distance to the sub-component connected to it in the next SiLaGO block to reduce power consumption in wiring \cite{paper5}. Hence this design can tackle the problem created by partial computation-centric customization. 

\par The SiLaGO is having hardware-centric implementation which allows the RTL, gates and physical levels to be hardware-customized. Also, in SOCs, only gates are customized, which gives chance for high power consumption in interconnects. SiLaGO, on the other hand, customizes RTL also, solving the problem created by inefficient software-centric implementation style \cite{paper5}. SiLaGO uses synchorocity principle to tackle the crucial problem created due to the lack of customization in dynamic runtime. Also spatial and temporal programming \cite{paper5} are used to handle this problem.

\subsubsection{Dynamically reconfigurable resource array (DRRA)}

Dynamically reconfigurable resource array (DRRA) is one of the optimum solutions in the field of coarse grain reconfigurable fabric that implements data which is streaming in a parallel manner for DSP application. This block is made of DRRA cells that are arranged in two rows and each one of them is composed of four components: register file (Reg. File), sequencer, data path unit (DPU), and two switch boxes (SWBs). The register file present in DRRA is operated using two read and two write ports to accept complex numbers with their address generation unit to perform spatial programming operation. DRRA sequencer acts as a configurable block to control and synchronize the statistical signal processing operations by programming switch box and DPUs. Calculations of DSP operations are performed using DPUs, ranging from circular convolution to IIR, FIR, scrambling, etc. Switch boxes configure the interconnects by controlling tri-stated multiplexers at the junctions.
    
\subsubsection{Distributed Memory Architecture (DiMArch)}
The DiMArch is a fabric dedicated to storage, which provides a large scratchpad memory and a parallel storage access, so as to match the high computation parallelism of the DRRA. DiMArch is also created by array disposition of SiLago cells; each cell takes the space of one or two DRRA rows so that it has enough room for a SRAM memory bank. Just like the RF, DiMArch cells are provided with AGUs that enable flexibility in address generation. A circuit-switched NOC glues the DiMArch banks together where the switches are considered to be programmed. In this way, different SRAM blocks can be clustered to make them look like one larger SRAM. The circuit switched NOC is preferable for data transfers because it's overhead is low in deterministic traffic patterns. DiMArch blocks do not contain their own sequencer for configuration, but they are handled by the ones inside the DRRA. This connection happens by a packet-switched configuration NOC, which has been chosen because it allows to easily reach any node in the network.

\section{State-of-the-art hardware architectures for feed forward deep neural network}
Artificial neural network (ANN), feed forward neural network (FFNN), or deep neural network (DNN) is one of the most promising algorithms for classification, detection, segmentation and recognition and a major part in the ever-increasing world of deep learning. The era of DNN, though started in 2006, became revolutionary with the research of Krizhevsky \textit{et al.} \cite{paper6}. FFNN is usually constructed by stacking multiple fully connected layers of neurons where the first layer serves as the input layer, last layer to predict the output and intermediate ``hidden'' layers to obtain the pattern hidden in the data to predict output. The weights of the interconnections between the layers are updated using backpropagation algorithm to minimize the loss at the output. Although extremely beneficial for prediction operation, the sheer number of weights, the exhaustive differentiation operation during backpropagation, presence of huge dataset and use of multiple training epochs make this procedure calculation-rich, bulky and hence should be made faster for user-benefit. To support the execution of DNN, the VLSI community need to accelerate their hardware infrastructure to provide aspects necessary for this computation.

\par
\begin{figure}[t]
    \centering
    \includegraphics[width = 9cm, height = 4.5cm]{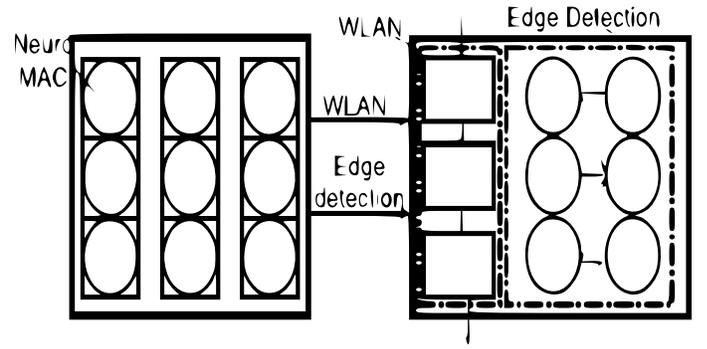}
    \caption{Block diagram of NeuroCGRA \cite{paper7}}
    \label{fig:neuro}
\end{figure}

 Jafri \textit{et al.} proposed a novel hardware architecture for deep learning support in \cite{paper7}. This hardware system, named ``NeuroCGRA'' is motivated by the better performance, low power consumption and low configuration area of course grain reconfigurable architecture (CGRA) compared to FPGA-based architectures. Also, CGRA has the ability to simultaneously host multiple operations in a single platform which enables the user to impose multiple deep neural network operations on a single CGRA platform to suit variability in performance without sacrificing the configuration area. The widespread application of neural network has already proved to be the state-of-the-art for estimation problem which can ensure better performance metrics and considers less space compared to the traditional implementation. However, the DSP and neural applications cannot be simultaneously provided by CGRA. ``NeuroCGRA'', the proposed architecture, on the contrary used a novel method to interleave the estimation operation replaced by neural computing using DNN. As in Fig. \ref{fig:neuro}, the part of NeuroCGRA can be dynamically altered to DSP and neural application depending on the application and hence exact or approximate calculations can be performed respectively. If the application requires the support of neural network, the platform is dynamically morphed to perform to approximate operation to increase the efficiency of the system. This architecture adds a new block to this already existing architecture, named translator and it generates different DRRA instructions. Memory files are used for storing the weights for every iteration to execute.

\begin{figure}[b]
    \centering
    \includegraphics[width = 9cm, height = 4.5cm]{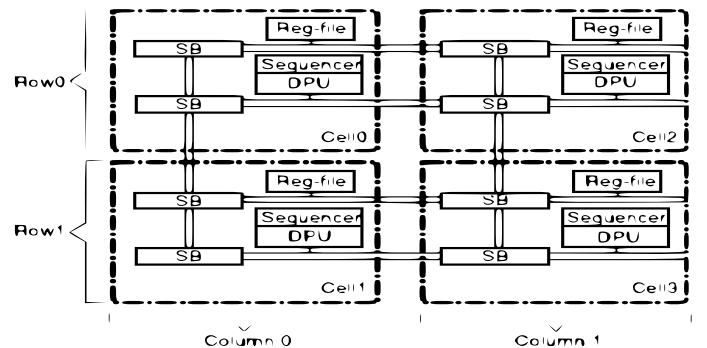}
    \caption{Neuron implementation using NeuroDPU \cite{paper8}}
    \label{fig:ngyen}
\end{figure}

\begin{table*}[t]
\centering
\caption{Comparison of performance for different state-of-the-art hardware architectures for DNN}
\begin{tabular}{|l|l|l|}
\hline
\multicolumn{1}{|c|}{\textbf{Authors}}                                              & \multicolumn{1}{l|}{\textbf{Contribution}}                                                                                                                                                                                                                                  & \textbf{Performance} \\ \hline
\multicolumn{1}{|c|}{Jafri \textit{et al.} \cite{paper7}}                                               & \multicolumn{1}{l|}{\begin{tabular}[l]{@{}l@{}}Introduced the method of morphable architecture to make neural network.\\The nodes present are interchangeably used for DSP and for DNN.\end{tabular}}                                                                    &     \begin{tabular}[c]{@{}l@{}}Negligible 4.4\% area,\\ 9.1\% power overhead\end{tabular}                 \\ \hline

\multicolumn{1}{|c|}{Ngyen\textit{et al.} \cite{paper8}}                                               & \multicolumn{1}{l|}{\begin{tabular}[l]{@{}l@{}}Proposed NeuroDPU block on the top of DSP  blocks for easy weight\\sharing and used Taylor series expansion to realize sigmoid and tanh.\end{tabular}}                                                                    &     \begin{tabular}[c]{@{}l@{}}Negligible 4.1\% area,\\ 8.8\% power overhead\end{tabular}                 \\ \hline

\multicolumn{1}{|c|}{\begin{tabular}[c]{@{}c@{}}Ardakani \\\textit{et al.} \cite{paper9}\end{tabular}} & \multicolumn{1}{l|}{\begin{tabular}[l]{@{}l@{}}Introduced integral stochastic computing for DNN implementation and\\adder and multipliers are approximated for making fast response.\end{tabular}}                                                                      &     \begin{tabular}[c]{@{}l@{}}Misclassification error\\ reduced from 4.7 to 2.3,\\ energy use from 2.96 to\\ 0.38, latency from 650 \\ to 30\end{tabular} \\ \hline    

Guan \textit{et al.} \cite{paper10}                                                                    & \begin{tabular}[c]{@{}l@{}}Divided the operation into high computation part and layer-specific part.\\ High-performance kernel was computed for the computation-intensive part\\\& communication bandwidth was optimized for the layer-specific part.\end{tabular} &         \begin{tabular}[c]{@{}l@{}}accuracy increases from\\ 89.9 to 96.6, energy use\\ reduces  to 0.12\end{tabular}             \\ \hline

\begin{tabular}[c]{@{}l@{}}Whatmough\\\textit{et al.} \cite{paper11}\end{tabular}                       & \begin{tabular}[l]{@{}l@{}}Introduced applications with a programmable accelerator design that\\ implements a powerful fully connected DNN classifier.\end{tabular}                                                                                                     &           \begin{tabular}[c]{@{}l@{}}98.36\% for MNIST data,\\ minimum energy 0.36 uJ\\ for accelerator\end{tabular}           \\ \hline

Hirtzlin \textit{et al.}  \cite{paper12}                                                                & \begin{tabular}[l]{@{}l@{}}Introduced resistive random access memories which are novel nonvolatile\\ memory technologies, which can be embedded at the core of CMOS \\and which could be ideal for the in-memory implementation of DNN.\end{tabular}              &     \begin{tabular}[c]{@{}l@{}}BER reduces to\\ $10^{-4}$ reduce RRAM \\ energy by 30 times\end{tabular}                  \\ \hline

Bai \textit{et al.}  \cite{paper13}                                                                     & \begin{tabular}[l]{@{}l@{}}Proposed hybrid structured DNN (hybrid-DNN) combining both spatial\\ and temporal deep  learning characteristics.\end{tabular}                                                                                                               &  \begin{tabular}[c]{@{}l@{}}99.03\% accuracy, 13.77\\ times reduction in error.\end{tabular}                    \\ \hline
\end{tabular}
\end{table*}

\par Ngyen \textit{et al.} proposed similar method to support deep learning activity as well as fundamental DSP applications in \cite{paper8}. Here the technique dictates a morphable way for block demonstration where different blocks like multipliers/ accumulators are arranged in a pipe lined fashion for producing the output estimates. 
The building blocks of a neural network are three elements, namely neurons, synapses and weights. The neuron model is followed while the calculation of weights in every neuron layer and and the results are calculated. These results can be used to extract features in a deep manner by passing it on to the next layer. Hence, in this way the output can be predicted from the classification layer. In order to realize and implement this on DRRA, one dedicated hardware is embedded with the CGRA which is termed as ``NeuroDPU''. This block is made to be connected with every DPU block of DRRA. This DPU, according to authors can be programming into normal mode, which is suitable for DSP operations and neuron mode, which does approximation on neural network. As in Fig. \ref{fig:neuro}, the NeuroDPU block mimics the neuron application working in a neuron mode by transmitting and receiving responses from associated layers using register files for three neurons and aggregating their values in a round robin fashion. This way, the result is generated and are stored in register. If the result is found to be greater than a predefined threshold, positive response is recorded.

\begin{figure}[b]
    \centering
    \includegraphics[width = 7cm, height = 4.5cm]{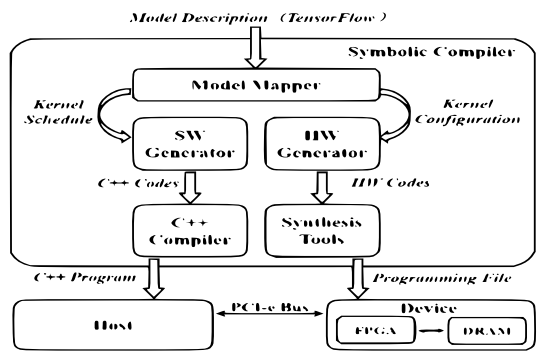}
    \caption{FP-DNN framework \cite{paper10}}
    \label{fig:guan}
\end{figure}
\par Ardakani \textit{et al.} proposed an efficient method to incorporate integral stochastic computing to estimate weights and responses in DNN \cite{paper9}. This method is used to solve the precision loss which crops up in adder because of the approximation operation and the delay happening because of high latency in the system. The delay happens because of high computation complexity and slow response. The method proposed in \cite{paper9} gives a novel approach to alleviate the latency problem better than the state-of-the-art. This paper primarily focused on DBN, hence for activation function, a novel tanh design is proposed using finite-state-machine. This method proved to be superior than the former stochastic implementation by giving praiseworthy performance by reducing the misclassification error. Also significant area coverage and latency reduction is observed, i.e. 45\% and 62\% respectively. The circuit is also synthesized by authors in 65 nm CMOS technology to estimate its performance in real scenario and it was found that the power consumption is reduced by 21\% compared to the conventional binary radix implementation without sacrificing the good performance.

\par Guan \textit{et al.} proposed a method to implement DNN in FPGA using RTL-HLS hybrid templates \cite{paper10}. Using this framework, the complex algorithms of DNN can be automatically mapped into the latest FPGA architectures for inference of models. This framework is completely automated and compared to the previous state-of-the-art design in accelerators, consider significantly less design time. In the same way, the activation functions are also designs which can be used by approximating the exponential operation for a shorter duration in Taylor series expansion method. The model-inference related operations are subdivided into two parts, out of which one part is computation-intensive, which performs high performance kernels and another part is termed as layer-specific part which performs small DSP operations, mostly multiplication and additions and are optimized by the bandwidth for communication. This FP-DNN framework is given in Fig. \ref{fig:guan} which accepts the tensor flow code and after analyzing the model description by model mapper, hardware optimization, kernel scheduling, host code generation and instantiating hybrid templates, the frameworks provides the output for FPGA application. This architecture can support all types of DNNs and can be extended for convolutional operations.

\begin{figure}[t]
    \centering
    \includegraphics[width = 9cm, height = 4.5cm]{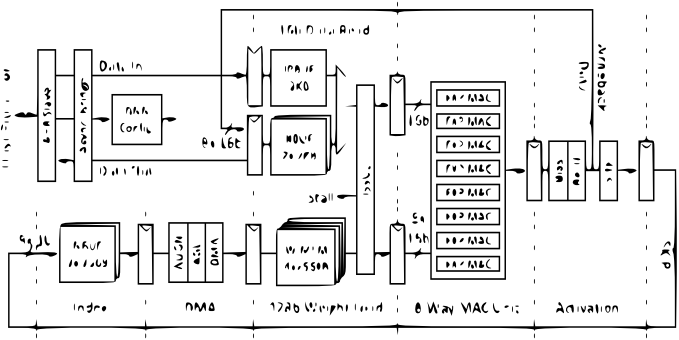}
    \caption{5-stage DPU MAC pipeline \cite{paper11}}
    \label{fig:whatmough}
\end{figure}

\par Whatmough \textit{et al.} proposed a novel design of SOC on 28 nm CMOS design for applications in internet-of-things (IoT). This model came in hand with an accelerator which can be programmed for customization and useful for implementation of a fully connected multi-layer perceptron algorithm with backpropagation operation \cite{paper11}. To incorporate this style of implementation, the DNN engine is considered to be programming machine for a five stage DNN by arbitrarily considering the values for weights and bias and storing them in a sparse matrix. This is useful to implement an arbitrary size of five-stage DNN with resource allocation scheme, as in Fig. \ref{fig:whatmough}. The basic of this architecture is constructed using an eight-way data path unit of multiply accumulate (MAC) unit. These lanes of these DPU can be used to perform the neural network operations and they constitute a single neuron to complete. Hence, at a time, only 8 weights can be generated. The weights are stored after the calculation of all neurons and hence long term memory is incorporated to accommodate all weight values inside the memory. To reduce the storage, however, weights are stored to be in either 8-bit or 16-bit signed type.

\par Hirtzlin \textit{et al.} investigated on performance and bit error range in binarized neural network using RRAM \cite{paper12}. Here the novelty of the work lies in the incorporation of a specific nonvolatile memory technology, which is named as ``resistive random access memory (RRAM)''. these memories are embedded at CMOS core and can be one of the best alternatives with power efficiency, performance precision and fast response to realize and implement deep neural network. The authors proposed this algorithm to be applicable for binary neural  network implementation which is considered to be a class of DNN with less memory usage. However, challenges crop up in this method when the devices are altered because this method cannot give reliable result in case of device variation. Investigation of authors dictates the impact of this model in bit error rate is profound, enhancing the performance by a significant amount. Also the error tolerance capability is better than the state-of-the-art, hence it can also be considered to be more reliable than others.

\par Bai \textit{et al.} proposed a neuromorphic computing based system to perform DNNs which can be considered to be a brain-like machine learning architecture. A DNN structured in a hybrid fashion is designed and implemented by authors in this work, where both spatial and temporal deep learning characteristics can be investigated and extracted. In this architecture, memristive synapses were used which works in a hierarchical manner. The memristor-based crossbar deploys feature extraction operation by unrolling the convolutional kernels into multiple single dimensional vectors, which upon operation forms the values pertaining to the columns of the crossbar. The prototype network was fabricated in ac 130 nm CMOS technology and from its experimental result, we can understand its brilliance in performance. Also it incurs parallelism with energy saving and low engineering cost which makes it ideal for implementation.

\par In TABLE I, these state-of-the-art hardware architectures are presented in a tabular format, where these architectures are described with respect to their contribution in the novelty of the literature and their performance for reducing misclassification error, memory, area and power overhead, fast convergence and high accuracy.

\section{State-of-the-art hardware architectures for softmax activation}
Activation function is an integral part for implementing neural network. These functions are nonlinear which helps the network to fit the distribution of the input dataset so that the newly generated model can predict the outcome of the unseen inputs. Generation of a activation function like tanh, ReLU or sigmoid are not computationally exhaustive. However, if we look back to softmax activation from other layers, we can see that this operation is much more rigorous and computationally expensive which depends on more than addition or multiplication operation. This block contains multiple exponential and division operations and hence making the design algorithm to be extremely difficult to make. Mainly it needs to combat two problems: the precision problem during division which can lead to classification error and overflow problem during the storage of huge number of exponential values. Hence, efficient implementation of softmax function has always been a promising research interest among VLSI scientists to improve latency and accuracy.

\begin{table*}[t]
\centering
\caption{Comparison of performance for different state-of-the-art hardware architectures for Softmax activation}
\begin{tabular}{|l|l|l|}
\hline
\textbf{Author}                                                        & \textbf{Contribution}                                                                                                                                                                                                                              & \textbf{Performance}                                                                                                 \\ \hline
Yuan \cite{paper14}                                                       & \begin{tabular}[l]{@{}l@{}}Performed logarithmic operation to avoid  the complicated division operation\\ into subtraction operation and completely remove the exponential operation\\ to avoid memory overflow.\end{tabular}               & \begin{tabular}[c]{@{}l@{}}Reduced critical path,\\ hardware complexity\\ and energy use.\end{tabular}               \\ \hline

Sun \textit{et al.} \cite{paper15}                                   & \begin{tabular}[l]{@{}l@{}}Here, exponentiation calculation of the softmax is split into several specific\\ basics which is implemented by ROM to simplified the hardware complexity\\ and logic propagation delay remarkably.\end{tabular}    & \begin{tabular}[c]{@{}l@{}}Power reduces from \\ 443 to 330 mW, clock\\ increases from 0.15 to\\ 1 GHz.\end{tabular} \\ \hline

Li \textit{et al.} \cite{paper16}                   & \begin{tabular}[l]{@{}l@{}}Used polynomial fitting of Taylor series expansion to accurately calculate the\\ value of the exponential then using logarithmic transformation to find out the\\ impact of each softmax function.\end{tabular} & \begin{tabular}[c]{@{}l@{}}Precision of magnitude\\ 10-5, clock increases\\ from 1 to 3.3 GHz.\end{tabular}          \\ \hline

\begin{tabular}[c]{@{}l@{}} Wang \\\textit{et al.} \cite{paper17} \end{tabular}     & \begin{tabular}[l]{@{}l@{}}Here exponential function is first is used on transform and then basic-split\\ operation is performed to increase its performance.\end{tabular}                                                                   & \begin{tabular}[c]{@{}l@{}}Power efficiency 463.04\\ from 330, area reduces \\ 24 times.\end{tabular}                \\ \hline
\begin{tabular}[c]{@{}l@{}} Kouretas \\\textit{et al.} \cite{paper18} \end{tabular} & \begin{tabular}[l]{@{}l@{}}Proposed hardware architecture was evaluated and synthesized in a 90-nm\\ 1.0 V CMOS standard-cell library.\end{tabular}                                                                                            & \begin{tabular}[c]{@{}l@{}}Area reduces 47\%,\\ energy consumption \\ reduces 43\%.\end{tabular}                     \\ \hline
\begin{tabular}[c]{@{}l@{}} Wang \\\textit{et al.} \cite{paper19} \end{tabular}     & \begin{tabular}[l]{@{}l@{}}This hardware design adds threshold layers to accelerate the training speed\\ and replace the Euler’s base value with a dynamic base value to improve\\ the network accuracy.\end{tabular}                         & \begin{tabular}[c]{@{}l@{}}Precision of magnitude \\ 10-6, accuracy increases\\ by 5\%.\end{tabular}                 \\ \hline
\end{tabular}
\end{table*}

\par
\begin{figure}[b]
    \centering
    \includegraphics[width = 9cm, height = 4cm]{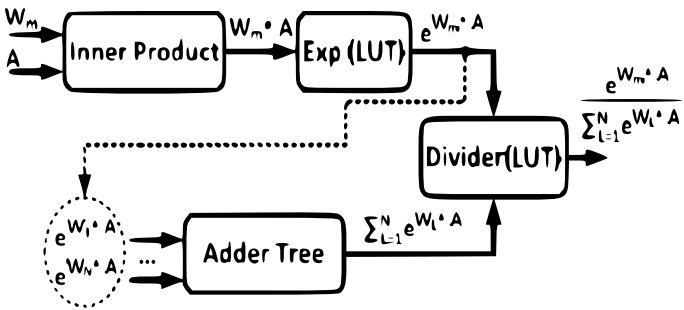}
    \caption{Flow diagram of hardware implementation of softmax activation \cite{paper14}}
    \label{fig:softmax}
\end{figure}

 B. Yuan addressed two major problems while designing softmax activation function in \cite{paper14}, i.e. 1) Softmax layer, unlike other activation function consumes a lot of parameters and values due to its computation of exponential and division operation, 2) This can cause overflow in memory due to vast calculation, precision problem because of large calculation and large critical path and more timing in comparison to other standard operations. He proposed to perform logarithmic transformation to avoid the complicated division operation into subtraction operation to boost the accuracy and completely remove the exponential operation to avoid memory overflow, as given in Fig. \ref{fig:softmax}. Also, he proposed to subtract the maximum value of the index from all to map the output value to range between 0 to 1 in order to further help reducing the memory overflow problem. Similar to this method, Sun \textit{et al.} proposed an algorithm to incorporate small look-up tables in \cite{paper15}. This mapping of values can be made straightforward using a lookup table. If we use this strategy, we can implement the complicated exponential calculation process very easily using multiple look-up tables. Also, as they are graded, they do not incur much memory overhead for performance and has proved to be actually better than the state-of-the-art in terms of memory. Hence, the exponential operation can be split open for individual ROMs which perform their calculation and portray their results and these values are multiplied. Instead of performing Taylor series expansion which gives approximate values, this can give accurate results in no time without sacrificing memory overhead. From experimental results, we can get that advantages like faster calculation speed, higher accuracy, larger throughput.

\par Li \textit{et al.} proposed an hardware architecture by considering different aspects of softmax generation \cite{paper16}. He used polynomial fitting of Taylor series expansion to accurately calculate the value of the exponential then using logarithmic transformation to find out the impact of each softmax function. Hence, it became a combination of lookup table and multi-segment linear fitting algorithm. Similar to this, for simplifying softmax function by Wang \textit{et al.}, authors have used mathematical transforms and linear fitting method \cite{paper17}. In this architecture, for optimization, different strategies are employed, e.g. reduction of strength of multiple algorithms, also addition methods are made faster. The exponential function is first is used on transform and then basic-split operation is performed to increase its performance. This block will incur approximation and ease of calculation without much sacrificing its accuracy.

\par Kouretas \textit{et al.} proposed an implementation methodology in hardware domain which can be used to generate softmax layer in CNN applications \cite{paper18}. In this state-of-the-art, several optimization techniques were carried out to accurately estimate the value of the softmax function. These approximated softmax function is then evaluated to incur better performance for hardware realization. 90 -nm CMOS technology was used then for fabrication of the design. The authors considered an approximation of equations which lead to a simplified expression. As we know, the softmax function for $z_{j}$ can be given as:
\begin{equation}
    f_{j}(z) = \frac{exp(z_{j})}{\sum _{k = 1}^{N}exp(z_{k})} \forall j \in [1,N]
\end{equation}
This equation, upon simplification using log transformation and maximum value subtraction, can be given as:
\begin{equation}
    log(f_{j}(z)) = z_{j} - (m + log(\sum _{k = 1}^{N}exp(z_{k}-m)))
\end{equation}

This value can be approximated to $z_{j} - m$ which is used for further calculation. Similar to this, Wang \textit{et al.} proposed another method with EEG signals \cite{paper19}. This version of algorithm for an optimized softmax layer implementation and its hardware designs are suitable for human emotion recognition algorithm while the input is EEG signal. the algorithm is considered to be CNN to extract spatial local features. Threshold layers are added with the hardware for speed acceleration and high performance curve fitting algorithm is employed to improve the accuracy of the architecture. Also batch normalization can be successfully implemented in a hardware-friendly way. TABLE II illustrates the comparison between these methods, their contribution and performance benefit.

\section{State-of-the-art hardware architectures for convolutional neural network}
The feed-forward deep neural network is nonetheless effective for extracting relationships between input and output. However, as time goes by, the shortcoming starts to crop up; mainly due to the massive memory storage and loss of local information. In neural network, all nodes in a layer are connected to all nodes in the next layers by dense connection and this uses a lot of memory to store these weights and biases. If we consider computer vision tasks, then the input is primarily image and video made up of lots of pixels. Hence if we try to incorporate dense connection, we will run out of storage. Another strikingly important feature of image frames is the presence of local spatial features, which can be extracted using spatial filtering in an efficient manner. However, conventional neural network tries to captures global information by flattening the images into a one-dimensional vector and loses the spatial information completely.

\par To alleviate these problems, convolutional neural network has been proposed. It incorporates weight sharing method to extract local spatial features like the filtering methods and usually it does not require any preprocessing. Weight sharing helps to minimize the memory requirement and calculation whereas the sparse connection helps to obtain local features to make it a locality-aware network for discriminative feature extraction. Convolutional neural network (CNN) is made up of three building blocks: convolutional layers for efficient feature extraction, pooling layers for feature aggregation and dimensionality reduction and fully connected layers for final classification. However, it has been experienced in \cite{paper6} that almost 90\% of the weights are present in fully connected layers, hence researchers have proposed many methodologies to avoid these layers as soon as possible.

\par Similar to this progress in algorithmic perspective, VLSI design engineers have also progressed to induce sparse connection and weight sharing to the conventional neural network to make it suitable for CNN applications. CNN does not contain many weights; however it considers huge calculation and parallel processing which have always become a daunting task for the hardware accelerator engineers to produce robust, low-latency and high-precision hardware architecture design. One of the earliest architectures in this flow is proposed by Qadeer \textit{et al.} \cite{paper20}. The architecture, named Convolution Engine (CE) reduces the overhead present in the register files due to huge memory content with the help of a FIFO-like structure. This provides the implementation of parallel feature extraction by shifting the data multiple times using the ability of this kind of storage structure. Using this data broadcast technique, a small register of 256-bits can accommodate sixty four ALUs and it saves energy as well. Further reduction happens in energy overhead by incorporating reduction tree to fuse multiple instructions together. Using this technique, memory overhead due to temporary storage can also be tackled. To aid two-dimensional feature extraction, the shift register allows itself to move across row and column, enabling for two-dimensional filtering mechanism. These architectural modifications enable this customized architecture to consume 8-50 times less energy compared to the state-of-the-art. Even compared to the most of the SIMD implementations, CE takes at least 8-15 times less energy.

\par In 2014, Chen \textit{et al.} proposed the DianNao architecture for deep CNN processing \cite{paper21}. In the naive and greedy deep neural network architecture, the hardware contains all synapses and nodes in the layout. Here the neurons and the synapses are implemented as logic circuits and RAM, respectively. This method which is conceptually recuperative towards the ideal implementation structure is capable of achieving energy efficiency because high speed and hence low energy can be obtained by this due to the small distance between intermediate nodes. However, for large Deep CNN, it cannot be used due to its size and scaling, because layout of large CNN models is very difficult, expensive and large. For example, here $16 \times 16$ layer considers less than $0.71mm^{2}$, however a $32 \times 32$ layer takes  more than $2.6mm^{2}$. Hence, a full layer of a single layer in the hardware would consist of large area. This accelerator consists of input buffer, output buffer and weight buffer for input neurons, output neurons and weights, respectively; connected with computation blocks such as neural function unit (NFU) and the control programs (CP). NFU performs arithmetic and logical operation and is used for classifier and convolutional layers. For sigmoid operation, authors used 16-bit fixed point arithmetic operation instead of 32-bit floating-point operation for low area and power consumption. All storage structures are split into sub-modules incorporate parallelization, low energy cost and latency. This hardware architecture has proved to be 118 times faster than the conventional SIMD processor and here energy overhead is reduced by 21.08 times. 

\par To incorporate pipeline, the DianNao architecture is made up of split buffers and NFU blocks perform computation in a parallel manner, where different operations are executed in subsequent stages in a hierarchical fashion, achieving better GPU performance in limited area and low power consumption. However, the shortcoming of this architecture occurs in convolutional and classifier layers due to memory bandwidth requirement. For these layers, required synapses are huge which is in terms of millions demanding high bandwidth. Also, the energy cost is increased by 10 times due to off-chip memory access. To combat this problem, Luo \textit{et al.} proposed DaDianNao architecture \cite{paper22}. This architecture is proposed to give solution for the machine-learning performance incorporating a system made up of multi-chip channels. As discussed, memory cost in DianNao is mainly due to weight sharing and high bandwidth is required for fetching the weights to their respective location. DaDianNao incorporates the following design modification to solve these problems: (1) The weights for the synapses are made to store closer to their destination neurons to minimize the data movement which saves both energy and time, (2) The architecture is made to asymmetric to make the nodes biased towards efficient storage pattern rather considering the computation, (3) Synapse values were frozen in their location and neuron values are transferred to reduce number of shifting values which reduces the bandwidth requirement by a significant margin, (4) The local storage structure is broken into multiple tiles to incorporate high internal bandwidth. This architecture reduces energy overhead by 185 times by increasing the speed over a GPU by 656 times. 

\par ShiDianNao, a machine learning hardware architecture for computer vision was proposed by Du \textit{et al.} \cite{paper23}. This energy-efficient architecture for computer vision can be used for real-time processing of image and video with any CMOS-based sensor. In this architecture, authors have focused on some specific factors, such as: (1) The images as a multi-dimensional tensor, (2) The weight sharing nature of the filters used for convolution operation. For this work, multiple operations are alleviated by authors, such as representing each neuron in the processing elements as a two-dimensional mesh and local distribution, receive, broadcast of kernel values and resulting feature map generation. The proposed architecture is made up of the following components: synapse buffer (SB), buffers incorporating neurons in input and output layers (NBin and NBout), neural function unit (NFU), arithmetic and logic unit (ALU) and instruction encoder (IE). The NFU blocks are used to perform fundamental neural network operations and ALU blocks are used to perform the activation operation. For sigmoid operation, authors used 16-bit fixed point arithmetic operation instead of 32-bit floating-point operation for low area and power consumption without compromising the accuracy. This architecture has proved to be faster than CPU by 46 times and GPU by 29 times. Also, in terms of the energy overhead, this architecture consumes 4688 times and 63 times less energy than GPU and DianNao, respectively. 

\par State-of-the-art deep CNN architectures are memory and computation exhaustive due to the presence of millions of parameters and equivalent range of calculations and hence are difficult to be embedded for efficient real-time applications. Although high-speed computation is supported using custom hardware, the required power is dominated by incorporating the weights to be fetched from DRAM, which consumes two times more energy than the conventional ALU operations. In 2016, Han \textit{et al.} introduced an efficient inference engine (EIE) architecture to combat this problem \cite{paper24}. This architecture is designed as an array of processing elements (PE) which stores parts of the SRAM elements. Each PE performs its corresponding memory related task to accomplish a part of the deep neural network. EIE is considered to be the first hardware accelerator incorporating sparse connectivity and weight sharing process. It works on on-chip SRAM which gives the benefit of 120 times less energy consumption than its counterpart involving the off-chip DRAM. It also exploits the sparsity in activation for minimizing number of computations. To avoid zero weights reference, it can save up to 70\% computation and 65\% energy in the state-of-the-art deep CNN algorithms. The sparsity in the matrix-vector multiplication is primarily applied in the dense layers which capture more than 90\% of the total connections, as stated in \cite{paper6}. Hence the reduction of computation and energy in this place has a synergistic effect on the overall performance. This architecture consumes 0.59W power compared to 15.57W in DaDianNao, As well as the area and energy efficiency for EIE is 6681 and 180606 respectively, compared to 2185 and 9263 respectively for DaDianNao.

\par The recently developed CNN models consist of huge storage for filter weights and on average 400k operation per input pixel. This situation challenges both the energy consumption and throughput of the processing accelerator. The total runtime is dominated by convolution operation because of its massive computation which might be an obstacle in the bandwidth requirement and high energy consumption. To avoid these problems, Chen \textit{et al.} proposed an efficient architecture for parallel data processing to minimize energy consumption with minimal data movement from the on-chip SRAM storage \cite{paper25}. The architecture, known as Eyeriss presents a novel dataflow, termed as row stationary (RS). This dataflow develops a spatial architecture and hence reduce data movement by exploiting reuse of filter weight values and the pixels of input tensors. This proposed dataflow is capable of adapting to different architectures for computer vision by utilizing the local storage as a processing engine, thereby reducing data storage as much as possible. This dataflow allows the design primitives to be updated for efficient 1D and 2D convolution operations in convolution layers and the fully connected layers are operated in the same manner without data reuse. To optimize energy cost, RS dataflow makes optimal data movement across all layers to maximize storage efficiency, ranging from global buffer to local register files. Experiments involving AlexNet architecture \cite{paper6} depict that this dataflow is more efficient in terms of energy consumption for both convolutional (2 times) and dense layers (1.3 times) than the state-of-the-art.

\par In these architectures, the performance was made to be optimum, however the low-power constraint has always become a pressing factor. The previous works also deployed modification regarding low-power consumption \cite{paper20}-\hspace{-0.1em}\cite{paper25} and less-memory access. The techniques followed by existing works are loop tiling, layer merging, parallelism, etc. However three limitations crop up amidst different modification, such as the nature of previous protocols to use one of the optimization techniques, hence losing the flexibility during design time. Also the limited access of compression for minimizing the memory access can become fatal, e.g. EIE applies it only in fully connected layers and Eyeriss does not apply it to kernels. Nonetheless, the requirement of a manual programming to map a CNN algorithm to the accelerator is also required. Jafri \textit{et al.} proposed MOCHA architecture to overcome these problems in 2017 \cite{paper26}. MOCHA applies reconfigurability to use the optimum protocol based on its application, making the architecture morphable. To combat the compression scenario, MOCHA applies compression with the flexibility of decompressing kernels and input images. Also, MOCHA applies an algorithm to automatically map a CNN architecture based on its need to the hardware for efficient accelerator implementation. Synthesis of the layout prevails that using MOCHA we can achieve energy efficiency of up to 63\%, throughput increases by 42\%, storage reduction happens by 30\%, while area requirement boosts up to 35\%, than the state-of-the-art. 

\par Another morphable accelerator, named ReCon was also proposed by Jafri \textit{et al.} \cite{paper27}. To obtain better accuracy for complex state-of-the-art CNN architectures, many accelerators are proposed for dedicated parallel processing and fast computation. While these architectures show promising result, they mostly suffer with two problems. One of them is the stagnant behavior of incorporating the same optimization algorithm for all layers irrespective of their needs which might slow down the performance, e.g. Eyeriss optimizes the convolution layers, however could not optimize the fully connected layers. Another problem in this case is the problem of compression which is generally not done throughout the layers. ReCon architecture incorporates morph-aware architecture to change optimization protocol depending on the needs and overcome the first problem. Also it introduces compression across all the layers including input layer to ascertain limited memory usage. ReCon memory is made of three sub-parts, consisting of a layer of off-chip DRAM holding input and kernel, a layer of on-chip SRAM working as a scratchpad memory and the register files to incorporate computations. Address generation units and memory clusters are applied to cascade multiple parallelism features used by SRAM scratchpad and register files. ReCon considers area overhead of around 30\%, however, it has 2 times more throughput than the state-of-the-art and provides best energy efficiency.

\par Chen \textit{et al.} introduced asynchronous behavior for CNN accelerator in 2018 \cite{paper28}. This accelerator design is made up of processing elements which are aligned as an array of $5\times5$ computation elements with the reconfigurable features for pooling and fully connected layers.  Different CNN models induce dynamic behavior for varying pool size and different pooling methods are also established by this. Local clock pulse signal replaces the global clock with the help of an asynchronous pipeline maintaining the preferable speed and accuracy. In the architecture, off-chip DRAM stores the input tensor values and information related to the dynamic configuration is written into the configuration array with the help of the controller. Input buffer helps the data to be read and register array stores the resultant tensors and the kernel weights. These values are supplied to the array for computation in convolution and pooling operation. Output buffer then stores the output again into DRAM. In the PE design, D flip-flops are used for local storage of values which are triggered by locally generated pulse signals. Each PE performs two operations such as the multiplication and the storing of operand values. Click signal enables flow of data by creating local pulse signal upon the arrival of request signal. The delays in each step are matched to maintain timing and critical path. Due to these reasons, asynchronous architecture works faster than synchronous pipeline. In the absence of request signal, the circuit stops performing to avoid unnecessary energy consumption. The performance of this architecture in LeNet-5 achieves 15.6\% increment in speed than the state-of-the-art without sacrificing the accuracy. 

\begin{table*}[t]
\centering
\caption{Comparison of performance for different state-of-the-art hardware architectures for CNN}
\begin{tabular}{|l|l|l|}
\hline
\multicolumn{1}{|c|}{\textbf{Authors}}       & \multicolumn{1}{c|}{\textbf{Contribution}}                                                                                                                                                                                                                              & \multicolumn{1}{c|}{\textbf{Performance}}                                                                                                                                                 \\ \hline
\multicolumn{1}{|c|}{Qadeer \textit{et al.} \cite{paper20}} & \begin{tabular}[l]{@{}l@{}}Parallel data extraction happens using shift registers in both directions. Instructions\\ are fused to avoid memory and energy overhead. Data are encoded to reduce computation. \end{tabular}                                           & \begin{tabular}[c]{@{}l@{}}Consumes 8-50 times\\ less energy compared\\ to state-of-the-art.\end{tabular}                                                                                 \\ \hline
Chen \textit{et al.} \cite{paper21}                         & \begin{tabular}[c]{@{}l@{}}Neural functional blocks are added as classifier and convolution layer. The operations are\\16-pt. to help low area and power consumption. Storage are split for parallelism.\end{tabular}                             & \begin{tabular}[c]{@{}l@{}}118 times faster and \\21.08 times less energy.\end{tabular}                                                     \\ \hline
Luo \textit{et al.} \cite{paper22}                         & \begin{tabular}[c]{@{}l@{}}Weights are stored closer to the application to minimize data movement and hence to\\ minimize time and power consumption. Synapses are not moved, instead neuron values\\ are moved to avoid memory overhead.\end{tabular}              & \begin{tabular}[c]{@{}l@{}}Reduces energy cost\\  by 185 times, speed \\increases 656 times.\end{tabular}                                                      \\ \hline
Du \textit{et al.} \cite{paper23}                           & \begin{tabular}[c]{@{}l@{}}Multidimensional tensor operation is performed considering it a 2-D mesh and 16-bit\\ arithmetic operation is done to limit power consumption.\end{tabular}                                                                                & \begin{tabular}[c]{@{}l@{}}Faster than CPU \& GPU\\ by 46 and 29 times. \end{tabular}    \\ \hline

Han \textit{et al.} \cite{paper24}         & \begin{tabular}[l]{@{}l@{}}Introduces sparse connectivity and encoding for proper weight sharing using on-chip\\ SRAM. Vectors are changed into matrix to operate in the processing elements.\end{tabular}                                         & \begin{tabular}[c]{@{}l@{}} 26.38 times less \\power than DaDianNao.\end{tabular}                               \\ \hline

Chen \textit{et al.} \cite{paper25}                         & \begin{tabular}[c]{@{}l@{}}Presents row-stationary dataflow to compute 1-D and 2-D convolutions. Optimal data\\ movement happens to minimize power consumption.\end{tabular}                                                                                         & \begin{tabular}[c]{@{}l@{}}The energy consumption\\ is 2 times lesser than EIE.\end{tabular}                                                                                              \\ \hline

Jafri \textit{et al.} \cite{paper26}                        & \begin{tabular}[c]{@{}l@{}}Introduces morphable architecture to choose best optimization protocol depending on \\application. To reduce memory usage, inputs and weights are compressed with easy technique. \end{tabular}                            & \begin{tabular}[c]{@{}l@{}}Energy, area, storage\\ reduces by 63, 35,30 \%.\end{tabular} \\ \hline

Jafri \textit{et al.} \cite{paper27}                  & \begin{tabular}[l]{@{}l@{}}Proposes a morphable architecture capable of obtaining best optimization protocol. The inputs\\ are clustered and then compressed to reduce the memory overhead. Sparse connection is also\\ reduced to avoid zero computation.\end{tabular}                                                              & \begin{tabular}[c]{@{}l@{}}Considers 30\% area \\ overhead, but throughput\\ is 2 times than MOCHA.\end{tabular}                           \\ \hline

Chen \textit{et al.} \cite{paper28}                   & \begin{tabular}[l]{@{}l@{}}Processing elements are arranged according to the kernel size. Reconfigurable features are\\ employed for pooling and fully connected layers also. The weights are stored closer to the\\ nodes to minimize energy consumption.\end{tabular}                                                              & \begin{tabular}[c]{@{}l@{}}It gives 15.6\% faster\\ response in LeNet-5\\ architecture.\end{tabular}                                                               \\ \hline

Sanchez \textit{et al.} \cite{paper29}                & \begin{tabular}[c]{@{}l@{}}Presents an optimization technique to convert 2-D convolution into 1-D convert to avoid the\\ transformation of vectors into matrices by using different vectors for the rows of the matrix\\ and then added to minimize energy consumption, computation and latency.\end{tabular}                        & \begin{tabular}[c]{@{}l@{}}Reduces the power \\ consumption by 7.3 times\\ without affecting accuracy.\end{tabular} \\ \hline

Li \textit{et al.} \cite{paper30}          & \begin{tabular}[l]{@{}l@{}}Uses tiling dataflow architecture, where the kernels are represented as tiles and portions of\\tensors are extracted for convolution, rather than taking the entire image and zero padding\\ which does not require vector-matrix transformation but reduces energy consumption and time.\end{tabular} & \begin{tabular}[c]{@{}l@{}}30 times faster and the\\ utilization efficiency is\\ maintained at 85\%.\end{tabular}                      \\ \hline

Kyriakos \textit{et al.} \cite{paper31}               & \begin{tabular}[l]{@{}l@{}}Divides the architecture corresponding to its software counterpart and uses mesh of processing\\elements to realize filter operation. Register banks are divided into sub-modules to implement\\parallel operation and pooling operation is performed using dedicated DSP blocks.\end{tabular}        & \begin{tabular}[c]{@{}l@{}}Increases the utilization\\ efficiency by 30\% than \\ the state-of-the-art.\end{tabular}                                               \\ \hline

Shang \textit{et al.} \cite{paper32}                  & \begin{tabular}[l]{@{}l@{}}Eliminates zero padding operation by creating coordinate relationship between subsequent\\layers. Also it develops better optimization strategy and uses data partitioning for parallelization.\end{tabular}                                                                                               & \begin{tabular}[c]{@{}l@{}}Provides 98.51\% and\\ 99.66\% computational\\ efficiency in AlexNet and\\ VGG-16, respectively.\end{tabular}                           \\ \hline

\end{tabular}
\end{table*}

\par Robust implementation of CNN accelerators rely on dynamically morphable architecture inherently bringing parallelism and efficient matrix vector computation. For optimization purpose, transformation of vectors into matrix form are also executed for make all multiplications similar. Though it eases the work flow, the negative aspect is the latency due to this transformation and redundant data generation which affects the memory. Utilization of GPU may ease these problems, however overall more computation, more memory burden and hence more power consumption and latency affect the performance of the accelerator for latency-sensitive and memory-exhaustive real-time applications. These problems mostly crop up in the first few layers where the number of pixels in the input tensor is large resulting in the massive computation and vector-to-matrix transformation. Sanchez \textit{et al.} demonstrates efficient hardware architecture to optimize original 2D convolution by incorporating the 1D architecture, primarily in the first few layers \cite{paper29}. In this design, authors argued to keep the kernel data in the on-chip SRAM and the reduction of this total amount is made to be possible by using 1D convolution kernels for the first few layers. For that, separate filters are used for both the dimensions. The feature maps are then concatenated by aggregating the produced results by reducing the memory utilization by a significant amount. This architecture reduces the power consumption by 7.3 times than the corresponding 2D design. 

\par Traditionally GPU was installed to incorporate parallelism for the implementation of CNN algorithm. However, as more complex architecture prevails the power budget and area constraint also increases leading to more size and power consumption. Hardware specialized CNN-based accelerators have already paved the way to overcome these problems. However, the state-of-the-art architectures either depend on one of the optimization structure or they are made to dynamically morph into one of the popular ones depending on their need. Li \textit{et al.} exploited one such dataflow for CNN application, i.e. tiling dataflow \cite{paper30}. However this dataflow does not possess better utilization efficiency in this context for the PEs getting larger than usual. To solve this drawback, parallelism is employed inside the hardware and a configurable hardware is demonstrated to incorporate different state-of-the-art CNN models by varying scheduling order which reduces hardware overhead to its best extent. The tiling dataflow is predominantly used for CNN accelerators which uses input buffer for storing feature maps, weight buffer for storing kernel values in the form of groups responsible for containing kernel values and convolution maps. To exploit the parallel operation used for CNN, kernel factors are unrolled to use several tiling work together. The processing elements contain the multiplication operation which is initiated using weight buffer and shift registers are used to contain the input feature values. Using this dataflow architecture, the accelerator outperforms the state-of-the-art by giving 30 times increment in speed and maintaining utilization efficiency at 85\%. 

\par The primary purpose of CNN is pattern recognition which is accomplished by performance-critical operation of filtering operation, where latency, area coverage and energy consumption plays a crucial role for incorporating it in real-time scenario. Kyriakos \textit{et al.} proposed an FPGA-based CNN architecture for efficient feature extraction in real-time with low latency, area and power budget \cite{paper31}. This FPGA-based protocol is subdivided into four fundamental blocks corresponding to its software counterparts: input layer, convolution layer, pooling layer and fully connected layer by experiencing parallelization features after division of arithmetic operations present in convolution layers into subsequent small modules and parallel computation using the processing element to increase throughput. The input layer receives the input from the off-chip DRAM access and then stores them into the on-chip SRAM by creating the appropriate window size for convolution operations in the subsequent layers. The images are stored in such a way so that by shifting redundant nature of CNN can be exploited. Parallel DSP blocks subdivide the convolutional operations in the subsequent layers, each containing separate filters for the application and the resultant is stored in SRAM memory. In the similar way, pooling operation is performed by incorporating DSP blocks across all the channels and dedicated array of blocks are assigned for this purpose. For fully connected layer, pipelined architecture is being followed with word-length optimization for finding tradeoff in throughput, latency, area and power consumption. This method increases the utilization efficiency by 30\% than the state-of-the-art.

\par For the implementation of CNN operation using kernel technique, zeros are padded for dimension matching which do not jeopardize with the accuracy, however consume a lot of unnecessary storage spaces which in turn increases the area, latency and power consumption. Also, during computation, if the number of pixels to be strided is less than the designated length of the FIFO structure where the input data is loaded, delay is incurred which in turn makes the system slow and useless for real-time applications. Also, filling zero execution happens in convolution layers for efficient feature extraction. However, it adds to extra area which is required to be added across all input tensors, also this extra zeros adds to the wastage of computational resources and memory exhaust problem specific for this application. Shang \textit{et al.} proposed LACS, a hardware accelerator for CNN to obtain high computational efficiency without sacrificing the accuracy \cite{paper32}. LACS eliminates the operation of zero padding and zero filling by incorporating coordinate relationship between different subsequent layers of the deep CNN architecture. Guided by a set of new instructions for efficient real-time CNN operation, it also develops better optimization strategy for obtaining the solution of stride length. A data-partitioning method is incorporated to obtain parallelism across data and for smooth running of the execution. The entire operation is controlled by executing the instruction stream between dependent modules for advanced control. LACS provide brilliant computation efficiency, e.g. 98.51\% for AlexNet and 99.66\% for VGG-16 and outperformed the state-of-the-art by a significant margin. The summary of these architectures are given in TABLE III with their corresponding novelty and performance analysis.

\section{State-of-the-art hardware architectures for generative adversarial networks}

Deep CNN architecture has considered to be the state-of-the-art for computer vision problem. However this is a data-hungry process and expects as much data as possible to extract all possible discriminative features necessary for classification purpose. Hence the big datasets are always in demand to ensure robust performance. Image augmentation techniques have guaranteed to alleviate this problem a bit; however it does not generate new data point, it helps to make CNN \textit{see} different representations of the same data and hence can become redundant and unnecessarily memory-intensive. Generative Adversarial Networks (GANs) are a set of unsupervised adversarial feature generation architecture proposed by Goodfellow \textit{et al.} which uses generative models to synthesize new images from the distribution of the existing training images \cite{paper33}. This architecture cleverly trains a generative model to implement and extract new data points which can be used for classification training, also it is widely used for many other applications.

\par GAN architectures are made up of two blocks: the generator block and the discriminator block, where generator block generates the fake/new data point and the discriminator discriminates them from the original point. The generator block starts with noise, preferably a 100-dimensional noise vector and this vector is passed through upsampling and de-convolution operation to  obtain the dimension of the training samples which are fed to the discriminator network working as a classifier to understand the authenticity of the generated image. The generator block wants the loss function to be small and the discriminator block wants it to be high. Hence, these two blocks play an adversarial mini-max game as proposed in Nash equilibrium and this process stops when the discriminator block no longer could identify the fake data points from the real one. Till then, both the models are trained to beat the other one, and after the execution, the newly generated fake data point can be considered a new data point available for training as it follows the same distribution and the fully trained discriminator block could not classify it to be a fake one. A variant of this vanilla GAN is proposed by Radford \textit{et al.} for deep convolution network with some additional constraints, DCGAN \cite{paper34}. 

\par GAN architectures constitute many complicated calculations due to its complex nature and the complexity is way more challenging than its counterpart in supervised learning technique. These operations include convolutions with various strides, transposed convolutions and the mini-max performance of the loss function and the combined training of the generator and the discriminator blocks. In conventional CNN training, one forward and one backward pass is required for training, however, on the other hand, GAN requires five such passes for each iteration. Also, the loss function must be synchronized with both the blocks which require immense hardware constraint for memory. To overcome these problems, Song \textit{et al.} proposed a micro-architectural solution for this unsupervised deep learning paradigm \cite{paper35}. To maintain the synchronization, value of the loss function are obtained from each individual data points in each epochs using a loop control statement by optimizing the result and resulting low memory usage. Also, as generator and discriminator blocks do not work together, time-multiplexing design is incorporated in order to optimize the space and reduce the idleness. Data flows are also maintained to optimize the computation overhead by skipping zero values and it also helps to reduce memory use. This architecture outperforms the conventional CPU and GPU by giving 8.3 times speedup and 6.2 times energy efficiency, respectively, without area overhead. 

\par
\begin{figure}[t]
    \centering
    \includegraphics[width = 9cm, height = 4.5cm]{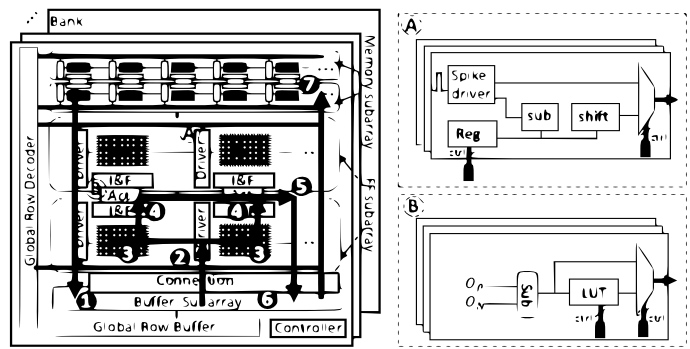}
    \caption{The ReGAN architecture \cite{paper36}}
    \label{fig:regan}
\end{figure}

 In GAN architecture, generator and adversarial blocks compete with each other in a mini-max game to achieve the loss function according to individual's wish and the trade-off point is when the discriminator block could not discriminate. This algorithm requires parallel execution of two massive blocks and hence high memory and power constraint comes into play the crucial role. Conventional CPU and GPU cannot perform this energy and memory-intensive process efficiently. In GAN, two networks are simultaneously trained and this training is difficult, also the generator path considers convolution operations with fractional strides and upsampling operations with different interpolation techniques. Chen \textit{et al.} proposed a ReRAM-based solution to alleviate these problems \cite{paper36}. ReRAM has considered being state-of-the-art for this purpose due to its fast response, high density, minimal power leak and better pipelining options. In this architecture, a pipelined protocol is followed for computing the convolution operations layer-wise and it does not require any extra processing elements. ReRAM buffers are created to store intermediate results and spatial parallelization is implemented with the sharing of due computation. This ReRAM-based hardware architecture for GAN depicted in Fig. \ref{fig:regan}, or termed as ReGAN is implemented on the platform and has shown to outperform the existing state-of-the-art performance and energy budget by 240 times and 94 times, respectively.  

\par In DNN architecture, local connecting buses interconnects the nodes, their memory and processing units which possess critical shortcomings such as latency, limited memory access and high power consumption. To tackle these problems, a novel processing-in-memory architecture has been devised as a potential solution for state-of-the-art accelerator design. This idea embeds logic states inside the memory for data processing in concurrent manner, thereby utilizing large bandwidth. The complex convolution operation can be converted into the simple addition and subtraction stages using ternary-GAN which not only reduces the area but also alleviates the problem of energy cost. Rakin \textit{et al.} introduced this processor-in-memory concept in ternary GAN \cite{paper37} for obtaining the advantages due to reduction of the access of main memory, power consumption reduction for in-memory computation, employment of parallelization to obtain fast and accurate response, etc. IN TGAN, the weights are ternarized to procure the least computation complexity and to maintain the bottleneck of power consumption. In this proposed architecture, authors achieved state-of-the-art performance in terms of 25.6 times better energy efficiency and 22 times increase in speed for GPU and this is 9.2 times more energy efficient and boosts up to 5.4 times than ReRAM accelerators.  

\par
\begin{figure}[b]
    \centering
    \includegraphics[width = 9cm, height = 3cm]{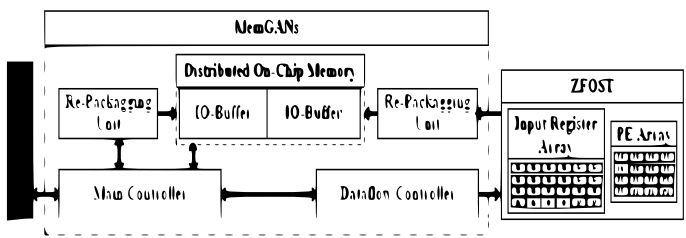}
    \caption{Block diagram of MemGANs \cite{paper38}}
    \label{fig:mem}
\end{figure}

 Apart from the conventional feed forward deep neural network, GAN is always different because of the nature of the existence of complex operations which involves fractionally-strided convolution and de-convolution operation makes the system clumsy, power and area constraint also prevails to abort for more complex loss function. Also the convolutional operations such as insertion of zeros and matrix vector conversion make the architecture very bulky. The on-chip conventional memory is also stored in a format which makes the requirement of the dataflow models inefficient. To obtain desired performance, the architecture needs a specialized memory access to provide input feature maps in a subtle way. Also, the ZFOST micro-architecture calls for a serious problem in orienting the feature maps during their reception from the off-chip DRAM memory. Hanif \textit{et al.} hence proposed MemGANs for efficient memory management for GANs in different micro-architectural paradigm. It helps to control the data prior to the storage depending on its pixel type \cite{paper38}. This architecture, given in Fig. \ref{fig:mem}, is composed of SPRAM, which is a single-port scratchpad memory for efficient memory allocation with the subdivision to demarcate proper storage places depending on the pixel values to meet the requirement of strided de-convolution. This proposed architecture enhances the performance of hardware accelerator for GAN by giving 3.65 times improvement in performance than the state-of-the-art. Also it reduces the read and write accesses by 85\% and 75\%, respectively. 

\begin{table*}[h]
\centering
\caption{Comparison of performance for different state-of-the-art hardware architectures for GAN}
\begin{tabular}{|l|l|l|}
\hline
\multicolumn{1}{|c|}{\textbf{Authors}} & \multicolumn{1}{c|}{\textbf{Contribution}}                                                                                                                                              & \multicolumn{1}{c|}{\textbf{Performance}}                                                                                                                         \\ \hline
Song \textit{et al.} \cite{paper35}                   & \begin{tabular}[l]{@{}l@{}}Develops loss function synchronization and uses time-division multiplexing to\\assign resources to generator and discriminator blocks.\end{tabular}                                                                                             & \begin{tabular}[c]{@{}l@{}}8.3 \& 6.2 times speed \& energy \\ efficiency.\end{tabular}                                        \\ \hline
Chen \textit{et al.} \cite{paper36}                  & \begin{tabular}[c]{@{}l@{}}Uses ReRAM-based architecture to solve the problem of fractionally strided\\ convolution and spatial parallelization is used for convolution.\end{tabular}                                                                                        & \begin{tabular}[c]{@{}l@{}}240 and 94 times performance\\ and energy efficiency.\end{tabular}                                                     \\ \hline
Rakin \textit{et al.} \cite{paper37}                 & \begin{tabular}[c]{@{}l@{}}Proposed processing-in-memory in ternary GAN to reduce power consumption\\ for in-memory computation and fast response.\end{tabular}                                                                                                             & \begin{tabular}[c]{@{}l@{}}Speed and energy efficiency\\ increases 22 and 25.6 times.\end{tabular}     \\ \hline

Hanif \textit{et al.} \cite{paper38}                 & \begin{tabular}[c]{@{}l@{}}Uses single-port scratchpad RAM for efficient memory allocation and puts the\\ data in the storage depending on pixel type.\end{tabular}                                                                                                           & \begin{tabular}[c]{@{}l@{}}Performance improved by\\ 3.65 times.\end{tabular}                                                             \\ \hline

Chen \textit{et al.}    \cite{paper39}              & \begin{tabular}[c]{@{}l@{}}Proposes to avoid zero padding for fractionally stride convolution to avoid\\upto 60\% calculation. The processing elements are made up of ReRAM \\blocks to increase speed.\end{tabular}                                                        & \begin{tabular}[c]{@{}l@{}}Performance improved by 23\\ times.\end{tabular}                                                                            \\ \hline

Di \textit{et al.} \cite{paper40}                    & \begin{tabular}[c]{@{}l@{}}Uses decomposition of transpose convolution to reduce computation and\\ avoids zero padding.\end{tabular}                                                                                                                                           & \begin{tabular}[c]{@{}l@{}}Processing performance\\ improves by 15 times.\end{tabular}                                                        \\ \hline

Chang \textit{et al.} \cite{paper41}                & \begin{tabular}[c]{@{}l@{}}Uses winograd algorithm to transform matrix to vectors for sparse connection.\end{tabular}                                                                                                                                              & \begin{tabular}[c]{@{}l@{}}Improves performance and speed\\ by 8.38 and 5.65 times.\end{tabular} \\ \hline

Roohi \textit{et al.} \cite{paper42}                 & \begin{tabular}[c]{@{}l@{}}Proposes approximate GAN architecture to get optimization in both hardware\\ and software using quantization to reduce computation. The approximation is\\\ also applied for deconv layers to avoid zeros for fractional strides.\end{tabular} & \begin{tabular}[c]{@{}l@{}}Increases speed and energy\\ efficiency by 35.5 \& 21 times.\end{tabular}                                                  \\ \hline
\end{tabular}
\end{table*}

\par Although GAN architectures have emerged to pave the ways for advanced research in computer vision; several factors impede the growth of the suitable hardware accelerator for GAN-based architecture. For example, the transposed and fractionally-strided convolution optimization is still an active research problem which cannot be implemented in a traditional hardware supporting DNN only. These operations involve the augmentation of low-dimensional feature maps into high-dimensional feature rich architecture for discrimination operation and involve insertion of zeros for this work which is memory, computation, energy and time-intensive. This constitutes to the severe under-utilization of resource due to almost 60\% involvement of multiplication and addition operations for this purpose. To alleviate these problems, Chen \textit{et al.} proposed ZARA, a dataflow accelerator which does not consider zero-padding for their execution made using ReRAM blocks \cite{paper39}. This method improves the efficiency in computation and avoids unnecessary utilization of memory resources by incorporating a mapper and operation scheduler following dataflow protocol for maximizing parallelism in both spatial and temporal domain. Vector-matrix multiplications are also effortlessly implemented without loss of area overhead using the ReRAM architecture, giving 100 times more efficiency than conventional design. ZARA observes the GAN model specifications involving network topologies, loss function optimization and parameter strain for both generator and discriminator block. Experimental results involving the implementation of ZARA using ReRAM chips define that the performance of the GAN-based accelerator improves by a factor of 23 over CMOS-based GAN implementation.  

\par During the computation in GAN architecture, the computation of transposed convolution in generator block is considered to be a performance-critical task which relies on several factor and hence creates a bottleneck in the optimization problem. This operation exerts the use of zero padding which results in unnecessary memory overhead and has proved to be attributing to more than 70\% invalid computation, which in turn increases energy budget, computation time and latency. Di \textit{et al.} explores an algorithm for hardware accelerator supporting GAN to obtain a resource-efficient framework \cite{paper40}. In this architecture, a novel dataflow is implemented for the processing of the transpose convolution by decomposition and rearrangement method which results in the praiseworthy reduction in computation complexity. This design structure incorporates the use of processing elements for successful implementation of multiplication and addition operations and for memory allocation, local buffers are utilized. This memory partition method invites parallelism into the system involving the reduction of computation time and better speedup. Parallelism incorporated is two-fold: the sequential parallel processing happens inside the processing units composing processing elements; also different processing units constitute parallelism by incorporating mutual connection. The processing units are composed of programmable logic resources which are dynamically morphed into the novel dataflow depending on the application. This architecture produces at least 15 times improvement in the processing performance than the state-of-the-art.  

\par FPGA-based designs are considered to be a promising factor to alleviate the problems of de-convolution layers with fractional strides and excessive zero padding due to their energy efficiency, good performance and reconfigurable nature. Winograd algorithm largely reduces the number of multipliers required for the computation and hence resource efficiency is improved. Chang \textit{et al.} proposed a hardware accelerator modification for de-convolution operation following Winograd algorithm, implementable on FPGA \cite{paper41}. This work accelerated the growth of the de-convolution layer related research for making it fast by means of converting it into a convolution later by using a spatial transform for improving throughput with parallelism without sacrificing the accuracy of the result. This algorithm is augmented with the Winograd algorithm to outperform the performance from the state-of-the-art. The architecture of de-convolution involving Winograd algorithm is composed of several intertwined processing elements and line buffers for inputs and outputs. The input logic fetches and performs the transformation with energy overhead. However, this overhead is mitigated by the incorporation of less zeros in the computation and hence less power consumption. The dataflow of this operation extracts the transformed features from the memory and proper reorganization is performed to mitigate the problem of unequal number of zeros inside the tiles. The sparse connection is exhibited for the easy operation and proper coding scheme is employed to minimize calculation. This accelerator outperforms the zero padded deconv and TDC-based deconv for being 8.38 times and 2.85 times faster. Winograd algorithm, on the top of this, boosts the performance by 7.5 times.  

\par For traditional computation, isolated units and the processing elements are aggregated via interconnects to expose the chance of high latency, huge memory burden, increased congestion, etc. The non-volatile memory technique exerts interesting features such as compatibility and high density integration. Roohi \textit{et al.} proposed a processing-in-memory based GAN architecture ApGAN or approximate generative adversarial network for optimization in both software and hardware perspective \cite{paper42}. In ApGAN, the architecture is designed for resource-limited environment, where both the hardware and software are optimized in order to minimize the overhead. The existing computations in GAN is modified in the convolution layers present in the generator and discriminator and replacing the multiplication operations with simpler addition and subtraction operations, which make them more efficient. Also, stringent measures are undertaken to quantize the selected portions in these blocks to reduce computation burden and memory resources without sacrificing significant loss in accuracy. The arithmetic unit is also replaced by an approximate form, primarily involving addition and subtraction operations and dataflow architecture is employed to incorporate parallelism in the accelerator design. Experimental results involving these structural and algorithmic modifications increase the speed by 35.5 times and energy efficiency by 21 times compared to the the performance obtained in the state-of-the-art in GPU platform. These architectures are also summarized in TABLE IV.

\section{Conclusion and future scope}
In recent world, the importance for automation in every aspect of our life has led researchers to pave the way for artificial intelligence, deep learning and high-end computing devices. While algorithms are getting stronger everyday, hardware obstacles are also being reduced by VLSI engineers to make these algorithms work seamlessly in an integrated manner providing outcomes with brilliant performance, high reliability and low latency. hence, in this way, the deep learning community and the VLSI frontiers are going hand-in-hand to ease the sheer load of mankind using automation, providing safety, security and certainty. The problem of devices which emanate heat can be resolved using dark silicon constraint, where all the transistors in the silicon surface are restrained to blow ``ON'' at a time, preventing accidental meltdown of devices when performance and execution is at peak. To provide the support required to maintain the dark silicon constraint, the heterogeneous multi-core processor and use of accelerators are established.
\par SiLaGO is a structure which can withstand this constraint and can be estimated to be the best alternative to tackle these ever-changing problems by mitigating use of interconnects and incorporating ``synchorocity'' in designs. General-purpose processors are meant to perform the DSP applications, many researches have been carried out to further accelerate the hardware architecture to perform deep neural network operation. Similarly, complicated, memory exhaustive and high-latency based activation function softmax is proposed by authors to ease the computation burden and memory overflow. The future generation research in this field lies in the further modification and implementation of superior power-saving architecture for multi-layer perceptron, restricted Boltzmann machine, recurrent neural network, convolutional neural network, reinforcement learning and so on. The SiLaGO architecture is now going through a surging stage from research to implementation. Some authors have already implemented SiLaGO as a neural network. Now weight sharing operation can be performed using SiLaGO module to calculate local spatial features to perform convolutional operation.

\end{document}